\documentclass[10pt,twocolumn,letterpaper]{article}

\usepackage{iccv}
\usepackage{times}
\usepackage{epsfig}
\usepackage{graphicx}
\usepackage{amsmath}
\usepackage{amssymb}

\usepackage[pagebackref=true,breaklinks=true,letterpaper=true,colorlinks,bookmarks=false]{hyperref}

\iccvfinalcopy % *** Uncomment this line for the final submission

\def\iccvPaperID{1151} % *** Enter the ICCV Paper ID here

\newcommand\changed[1]{\color{black}#1}

\newcommand\blfootnote[1]{%
  \begingroup
  \renewcommand\thefootnote{}\footnote{#1}%
  \addtocounter{footnote}{-1}%
  \endgroup
}

\newcommand\sota{state-of-the-art }

\ificcvfinal\fi
\begin{document}

%%%%%%%%% TITLE
\title{Cascaded Sparse Spatial Bins for Efficient and Effective Generic Object Detection}

%\author{David Novotny\'}

\author{David Novotny\textsuperscript{1,2}\\
\textsuperscript{1}Visual Geometry Group\\
University of Oxford\\
{\tt\small david@robots.ox.ac.uk}
\and
Jiri Matas\textsuperscript{2}\\
\textsuperscript{2}Center for Machine Perception\\
Czech Technical University in Prague\\
{\tt\small matas@cmp.felk.cvut.cz}
}
\maketitle

\begin{abstract}
%novel method
%window scoring
%3 descriptors - novel BEV, cnn spp + edgeboxes
% - utilizes spatial bins
%    -> feature selection to speed up
%
%performance part ...

A novel efficient method for extraction of object proposals is introduced.
Its "objectness" function exploits deep spatial pyramid features, a novel fast-to-compute HoG-based edge statistic  and the EdgeBoxes score \cite{zitnick2014edge}. The efficiency is achieved by the use of spatial bins in {\changed a novel} combination with sparsity-inducing group normalized SVM.

\mbox{State-of-the-art} recall performance is achieved on Pascal VOC07, significantly outperforming methods with comparable
speed. Interestingly, when only 100 proposals per image are considered the method attains 78\% recall on VOC07.
The method improves mAP of the RCNN \sota class-specific detector,
increasing it by 10 points when only 50 proposals are used in each image.
{\changed The system trained on twenty classes performs well on the two hundred class ILSVRC2013 set confirming generalization capability.}

\end{abstract}

%%%%%%%%% BODY TEXT
\section{Introduction}

{\changed Object detectors have often been applied in the sliding window fashion scoring bounding boxes in all considered positions, scales and aspect ratios using either an inexpensive classifier 
\cite{felzenszwalb2010object,dalal2005histograms} or cascades \cite{viola2001rapid,vedaldi2009multiple}.  
The development of sophisticated and computationally demanding 
deep learning based object detectors \cite{girshick2014rich,he2014spatial} 
 stressed} the need to decrease the number of fully scored bounding boxes while retaining high recall levels. 
\blfootnote{
The authors were supported by the Czech Science Foundation project GACR P103/12/G084 and by the Technology Agency of the Czech Republic TE01020415 V3C -- Visual Computing Competence Center.}

Similar to the first stages of the cascades,
object proposals \cite{carreira2010constrained,alexe2010object,van2011segmentation}
are class-agnostic high-recall-low-precision object detectors that tackle computational efficiency by rejecting likely background regions while retaining bounding boxes covering instances of the semantic object classes 
which are later classified by the final class-specific object detector. 

%the option of trying all possible image subwindows.
%State-of-the-art class-specific object detectors have too high 
%computational demands and so cannot be applied in the sliding window fashion.
%Thus a suitable set of bounding box proposals is typically required to make the detection
%times feasible.

State-of-the-art proposal methods either generate candidate boxes from image segments, e.g. groups of superpixels or randomly initialized binary segmentation outputs \cite{van2011segmentation,manen2013prime,carreira2010constrained,endres2010category,arbelaez2014multiscale}, or
select proposals from a large pool of densely sampled image regions 
according to a predefined "objectness" score \cite{alexe2010object,rahtu2011learning,cheng2014bing,zhang2011proposal}. 
The latter approaches, also known as "window scoring" methods \cite{hosang2015what}, utilize diverse types of inexpensive features that most commonly capture edge statistics along the scored region boundaries \cite{rahtu2011learning,zitnick2014edge,cheng2014bing}. 
% pick proposals according to a score of a set of descriptors extracted from
% densely sampled image regions \cite{alexe2010object,rahtu2011learning,cheng2014bing,zhang2011proposal}. 
% The latter approaches also known as "window scoring" methods \cite{hosang2015what} utilize different types of cheap features that mostly capture edge statistics along the scored region boundaries \cite{rahtu2011learning,zitnick2014edge,cheng2014bing}. 

%In this paper, we introduce a novel method for extraction of object proposals using the window scoring approach.
%More precisely, we demonstrate how to use the powerful CNN features to boost the performance of class-agnostic object detectors
%while keeping the computational demands of the proposed method at a tolerable level. 
%While utilizing cheap proposal techniques from \cite{zitnick2014edge} and \cite{van2011segmentation} 
%we exploit the deep features introduced in \cite{he2014spatial}. In addition we employed another fast-to-compute feature in combination with the EdgeBox score \cite{zitnick2014edge} to further improve the performance of the proposed approach. Our method attains low execution times mainly due to the employment of a sparse SVM based feature selection.

In this paper we introduce a method for extraction of object proposals using the window scoring approach.
The key novelty is the use of spatial bins \cite{lazebnik2006beyond} in combination with group normalized SVM which 
enables to carry out the superficially complex proposal score computation surprisingly fast. 
The proposed objectness function exploits the following sources of information: the deep spatial pyramid features introduced in \cite{he2014spatial}, a novel fast-to-compute HoG-based edge statistic which also takes advantage of the spatial bins and the EdgeBoxes score \cite{zitnick2014edge}. Optionally, recall of the method can be boosted by selective search \cite{van2011segmentation} but this slows down the detection slightly.

We experimentally verified that:
(1) The introduced method gives \sota results when comparing the overlap-recall curves. (2) The performance of the \sota class-specific RCNN detector \cite{girshick2014rich} on our object proposals improves and the performance is less sensitive to the number of used proposals in comparison with other \sota proposal methods. (3) Despite being trained on a dataset that contains a small set of distinct object classes, it generalizes to previously unseen classes. These factors result in a proposal method that is as fast as standardly used Selective Search in "fast mode" \cite{girshick2014rich,he2014spatial} while achieving better recalls. 

The rest of the paper is organized as follows. Sect.~\ref{sect:related} gives brief information about modern proposal approaches. A concise explanation of our method is provided in Sect. \ref{sect:overview}. The details about the features we use are in sections \ref{sect:ebfeat}, \ref{sect:cnnsppfeat}, \ref{sect:bevfeat}.
An explanation of the utilized feature selection approach resides in Sect. \ref{sect:binSparsity}. Sect. \ref{sect:arnms} explains the special type of non-maximum suppression we employ and Sect. \ref{sect:experiments} provides results and discussions of concluded experiments. Sect. \ref{sect:conclusion} presents conclusions of our work.

\section{Related work}                     
\label{sect:related}

Noting that an exhaustive description and evaluation of recent \sota is presented in Hosang \etal \cite{hosang2014good,hosang2015what}
a brief explanation of key proposal methods is given in this section.

Many proposal methods build on the seminal Selective Search (SS) of Van de Sande \etal \cite{van2011segmentation} which progressively aggregates superpixels obtained by the Felzenszwalb and Huttenlocher method \cite{felzenszwalb2004efficient} into larger groups based on their similarity. The SS approach still is one of the best in terms of recall and quality of the proposal localization when a large number of candidate windows is requested  (more than 1000 per image). Its disadvantage is the inability to select a  smaller convenient subset of candidates since it lacks a suitable way of evaluating proposal importance. The relatively slow extraction speed of 10 seconds per image is improved in the "fast mode", accelerating  to $\sim$2.5sec/image. However, the accelerated mode looses the high recalls when larger proposal pools are requested. Modifications of Selective Search include Randomized Prim's \cite{manen2013prime} which learns superpixel similarity measures and employs an order of magnitude faster grouping algorithm. However this comes at the cost of lower attained recalls. 

%\cite{manen2013prime} is resemblant to \cite{van2011segmentation}, since it also
%merges superpixels, however learned superpixel similarity measures are utilized instead of the hand-coded ones. Furthermore a randomized spanning-tree algorithm is used for finding the groups of superpixels. 

In Constrained parametric min-cuts \cite{carreira2010constrained} (CPMC), every proposal is a solution of a min-cut segmentation problem initialized with a random seed. The proposals are ranked on the basis of various types of features. While this approach is able to deliver \sota recall and localization performance, its speed of a few minutes per image is a significant disadvantage.
The approach of Endres and Hoiem \cite{endres2010category} bears resemblance to 
CPMC in the sense that a foreground / background regressor initialized by different seeds is learned for obtaining a set of proposals that are subsequently ranked. The method is slow,  about two times faster than CPMC.   
Multiscale Combinatorial Grouping \cite{arbelaez2014multiscale} (MCG) introduced a fast hierarchical segmentation algorithm. On top of that, an efficient exploration of the large combinatorial space of the produced segments is employed in the grouping stage. While the method achieves \sota performance in terms of recalls it is slow at approx. 30 sec per image.

Rigor \cite{humayun2014rigor} address the speed problem  of CPMC by reusing max-flow computations. Similarly, Geodesic object proposals \cite{krahenbuhl2014geodesic} replace the min-cut algorithm with a  much faster geodesic distance transform seeded by a learned foreground/background regressor. While Rigor has the same speed as Selective Search it has slightly lower recalls. Geodesic proposals run at 1 image/sec and their recall is comparable to Selective Search. However, due to its inability to assign scores to proposals, it is {\changed it is not obvious how} to limit the number of  output candidates. 

Rantalankila \etal \cite{rantalankila2014generating} combine the superpixel merging approach \cite{van2011segmentation} with CPMC \cite{carreira2010constrained}. The results in \cite{hosang2015what} indicate that the method is inferior to \sota both in terms of speed and attained recalls. 

%Endres \etal \cite{endres2010category} bear resemblance to 
%CPMC in the sense that again a foreground / background regressor initialized by different seeds is learned for obtaining initial set of proposals. After that a ranker that utilizes a large variety of features is used to sort the regions according to their probability of containing an object. 

%RIGOR \cite{humayun2014rigor} tackles the biggest downside of CPMC which is its slowness by reusing max-flow computations between graphs while Geodesic object proposals \cite{krahenbuhl2014geodesic} replace the min-cut algorithm with much faster geodesic distance transform seeded by a learned foreground/background regressor. 

%Rantalankila \etal \cite{rantalankila2014generating} combine the superpixel merging approach \cite{van2011segmentation} with CPMC \cite{carreira2010constrained}. It consists of two stages - local and global. Instead of running CPMC on top of raw pixels, they opt for grouping superpixels during the global stage of their algorithm. The local stage is very similar to \cite{van2011segmentation}, the difference lies in the set of features used for calculating the superpixel similarity.

Methods based on the sliding window paradigm extract features lying inside predefined bounding boxes and score them using a learned classifier. The work of Alexe \etal \cite{alexe2010object} was the first of this kind. Later Rahtu \etal \cite{rahtu2011learning} improved \cite{alexe2010object} by adding more powerful features and by learning a more convenient cascade of structured output SVM classifiers \cite{tsochantaridis2004support}. Additionally, Zhang \etal \cite{zhang2011proposal} proposed cascade of ranking SVMs that score inexpensive edge-based features.
Despite the high speed of these approaches their recall performance is inferior to \sota \cite{hosang2015what}.

EdgeBoxes (EB) is a fast proposal algorithm, running at 0.3 sec per image, with compelling performance \cite{zitnick2014edge}. EB scores proposals using a single feature - the number of contours that are fully enclosed by a bounding box minus those that overlap its boundary. After scoring each region, non-maximum suppression (NMS) takes place. Different overlap thresholds of NMS provide a compromise between accuracy and recall. 

%Authors show that different overlap thresholds of NMS provide different compromises between the accuracy of their proposals and their recall. 

BING (\cite{cheng2014bing}) is also based on edge features and provides fairly high recall at low IoU\footnote{IoU: Intersection Over Union Pascal bounding box overlap metric.} thresholds at the speed of 300 frames per second. However, its performance is significantly inferior to other methods at higher IoU thresholds. This leads to poor performance when used in combination with class-specific object detectors \cite{hosang2014good}. Moreover, its high recall is more a result of the careful placement of initial bounding boxes than of the discriminative power of the used features and classifier \cite{zhaocracking}.

Deep learning methods have recently entered the field of generic object detectors. DeepMultiBox \cite{erhan2014scalable} directly regresses the locations of proposals from an image using a deep convolutional network. Szegedy \etal \cite{szegedy2014scalable}
builds on top of \cite{erhan2014scalable} and achieves \sota detection performance on ILSVRC2012 \cite{ILSVRCarxiv14}.
Although both \cite{erhan2014scalable} and \cite{szegedy2014scalable} evaluate the performance of a class-specific detector that uses their proposals, neither paper presents overlap-recall curves of their generic object detectors preventing comparison with other \sota proposal methods. 

A very recent work of Karianakis \etal \cite{karianakis2015boosting} uses integral channel features detector \cite{dollar2009integral}. The individual channels are filters from the convolutional layers of a deep neural network. This work is perhaps the most similar to our approach. 
The differences include: (1)~the way our deep features are extracted and how the feature selection is carried out. (2)~Besides deep features, we use a novel edge-based statistic. (3)~We use SVM classifier instead of AdaBoost. (4)~Our results are superior in terms of overlap-recall curves.

%
%The most important difference lies in the way our deep features are extracted and how the feature selection is carried out. Furthermore our work proposes to use another different type of features in combination with SVM classifier instead of boosting. In addition, our method provides better results in terms of overlap-recall curves.

%{\red
%To conclude there is a small number of approaches that make use of the deep convolutional features, which recently became very popular due to their superior performance on both class-specific image classification and object detection tasks \cite{krizhevsky2012imagenet,girshick2014rich,sermanet2013overfeat,he2014spatial}. Although very recently three proposal methods that utilize deep features emerged - \cite{erhan2014scalable,szegedy2014scalable} and \cite{karianakis2015boosting} - the first two works provide only recalls achieved at 50~\% IoU\footnote{Intersection Over Union Pascal metric for evaluation of bounding box overlap.} threshold which was shown to be an inconvenient way of evaluating quality of object proposal methods (due to the "Achilles' heel of \#DR-\#WIN evaluation framework" \cite{zhaocracking}) 
%and the last one has performance inferior to existing \sota methods at higher IoU thresholds (class-specific object detectors used in combination with proposal methods that have lower recall at higher IoU thresholds typically exhibit poor performance \cite{hosang2015what,hosang2014good}.}

\section{Method overview}

\begin{figure*}[t]
\begin{center}
   \includegraphics[width=1\linewidth]{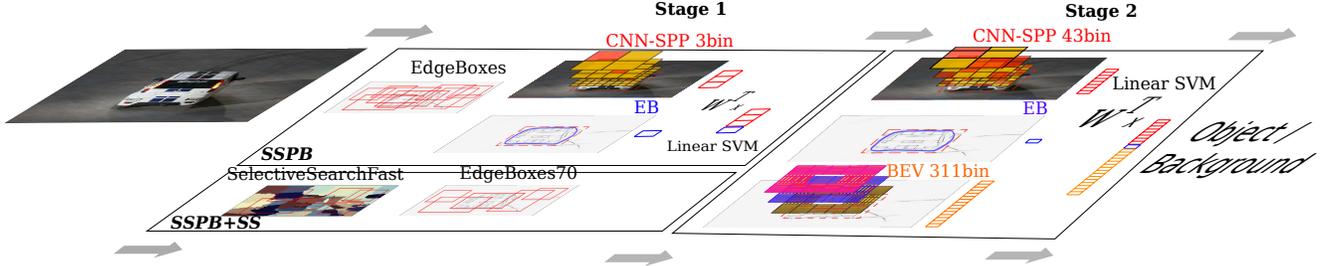}
\end{center}
   \caption{\textbf{An overview of our method.} The first stage of the cascaded approach consists of either extracting EdgeBoxes70 together with Selective Search (fast mode) proposals (SSPB+SS) or filtering a large input set of dense EdgeBoxes proposals using SVM that utilizes fast \mbox{CNN-SPP} features (SSPB). During the second stage three descriptor types are extracted from each window and scored by a linear SVM to obtain the final objectness score. }
\label{fig:overview}
\end{figure*}

%Because we wanted to keep the computational demands of our method as low as possible when choosing between the segmentation and window scoring approaches we opted for the latter one

We selected a window scoring approach since the segmentation based ones are, apart from Selective Search in "fast mode", 
very slow due to  their reliance on superpixel generation or min-cut segmentation algorithms.
Since most of the large pool of tested image windows contains background, we employ a well-known paradigm consisting of a cascade of progressively more complex classifiers \cite{viola2001rapid,vedaldi2009multiple} to introduce an early rejection mechanism.
While there are many possible choices for the types of classifiers in the cascade, we utilize binary linear SVM due to its high speed%
\footnote{Test showed that the usual choice of AdaBoost \cite{schapire1999improved} is inferior.}.

%Also, the group-normalized SVM based feature selection objective function largely resembles the one of the standard SVM solver, most likely producing features specifically tailored for the final binary SVM classification problem.

The first stage of the cascade reduces the initial number of $\sim$100k of all considered windows roughly by a factor of 10. During the second stage a linear SVM classifier produces final window scores on the basis of computationally more expensive features. The last step consists of a special type of non-maximum suppression (NMS) termed ARNMS that optimizes average recall (AR)\footnote{The area under the recall-overlap curve evaluated for IoU thresholds ranging from $0.5$ to $1$ \cite{hosang2015what}}. Details about ARNMS are provided in Section \ref{sect:arnms}.
%ability to cope with various correlations in the input data (in contrast to the standard choice of Adaboost \cite{schapire1999improved} which considers one feature at a time). 
The features that describe each bounding box are:

\textbf{CNN-SPP:}
We follow up on the success of convolutional neural networks on the both object detection and image categorization  \cite{krizhevsky2012imagenet,girshick2014rich,sermanet2013overfeat,he2014spatial} and use them as our primary bounding box descriptor. To maintain computation efficiency and thus high speed,  we employ the fast deep feature extraction technique from \cite{he2014spatial} that is able to process several thousand bounding boxes per second.
%deep feature extraction technique from \cite{he2014spatial}.

\textbf{BEV:} (stage 2 only) Since various edge statistics are a useful objectness cue \cite{rahtu2011learning,alexe2010object,zitnick2014edge} we introduce a novel Boundary Edge Vector feature (BEV) inspired by the Boundary Edge distribution introduced in \cite{rahtu2011learning}. 

\textbf{EB:} Due to an immense speed of the extraction of the EdgeBoxes score \cite{zitnick2014edge}, we include it as another type of an edge statistic feature.

\label{sect:overview}

%We designed our objectness pipeline as a cascade of classifiers. Our system consists of two stages. During each stage a descriptor is extracted from input regions and scored using a binary linear SVM classifier. We make use of three types of descriptors:
%%\begin{description}
%%\item [CNN-SPP] Max-pooled CNN convolutional filters as proposed by \cite{he2014spatial}.
%%\item [IBE] Improved Boundary Edge distribution feature first introduced in \cite{rahtu2011learning}.
%%\item [EB] The score that EdgeBoxes \cite{zitnick2014edge} assign to the produced proposals.
%%\end{description}
%
%\textbf{CNN-SPP:} Max-pooled CNN convolutional filters as proposed by \cite{he2014spatial}.
%
%\textbf{IBE:} Improved Boundary Edge distribution feature first introduced in \cite{rahtu2011learning}.
%
%\textbf{EB:} The score that EdgeBoxes \cite{zitnick2014edge} assign to the produced proposals.

Additionally, we \emph{speed up extraction of BEV and CNN-SPP features by employing a group-normalized SVM based feature selection algorithm} that automatically collects the set of spatial bins which are the most important for the final classification decision; details are provided in Section \ref{sect:binSparsity}.

A schematic illustration of our method is presented in Figure \ref{fig:overview}. In what follows, both classifier stages are discussed in detail.
An in-depth explanation of the aforementioned features is provided in sections \ref{sect:ebfeat}, \ref{sect:cnnsppfeat}, \ref{sect:bevfeat}. 

\subsection{Classifier cascade: Stage one}

We propose two ways of producing an initial set of regions during the first stage either of which can be used.

%The later stages of the pipeline utilize solely one of these two options:

\textbf{Selective Search + EdgeBoxes70}: Selective Search (SS) \cite{van2011segmentation} regions (using its "fast mode") 
merged with EdgeBoxes70 (EB70) \cite{zitnick2014edge} bounding boxes.

%\item [EdgeBoxes only]
\textbf{EdgeBoxes only}:
Due to the relative slowness of the Selective Search proposals in comparison with other parts of our pipeline, the second initialization type employs only EdgeBoxes. We set its $\alpha$ parameter controlling the density of the bounding box sampling to a relatively high value of $0.75$ to force the generation of an overcomplete pool of regions. EdgeBoxes $\beta$ parameter was set to $1$ effectively removing the non-maximum suppression step. This setting produces around 50k regions in 0.5 seconds per image.
%Subsequently, the size of this initial set is decreased by taking 
Subsequently, we take only 30k highest scoring regions according to the EB score. On this set, CNN-SPP descriptors are extracted and appended to the EB scores to form a final stage-1 descriptor which is later scored by an SVM. {\changed The ARNMS based on the SVM scores reduces} the output to the desired number of 10k boxes.

The second stage is common to both types of stage-one initialization. Two independent versions of our approach can be considered depending only on the chosen initialization type. We term the pipeline that is initialized by EB70 in combination with SS proposals \textbf{SSPB+SS} and the pipeline that utilizes EdgeBoxes only \textbf{SSPB}.

\subsection{Classifier cascade: Stage 2}

%The second stage of the cascade is common to both SSPB and SSPB+SS pipelines.
A fixed length descriptor, consisting of the three types of features (EB, BEV, CNN-SPP - described in sections \ref{sect:ebfeat}, \ref{sect:cnnsppfeat}, \ref{sect:bevfeat}) 
concatenated into a single vector
is extracted from each of the 10k bounding boxes and scored using a fast linear binary SVM. The maximum number of 10k input bounding boxes was experimentally found to give good trade-off between the speed and recall. ARNMS which is specifically tuned for the amount of requested proposals is the final step of our method. 

% \section{Features explanation}

% \label{sect:usedFeats}

% In this section we explain the three types of features that are used as bounding box descriptors. 

\section{EdgeBoxes feature (EB)}

\label{sect:ebfeat}

The fastest feature type is the contour score that EdgeBoxes assign
to its proposals. 
%We modified the publicly available EdgeBoxes code to be able to rapidly extract 
%its edge affinity score from the set of predefined bounding boxes. 
Note that in the case of SSPB+SS, besides retaining the score of the extracted EdgeBoxes70, we further use the publicly available EdgeBoxes
code to obtain the EB score of the additional Selective Search proposals (without performing the region refinement).

\section{CNN-SPP feature}

\label{sect:cnnsppfeat}

%maxpool, CNN, spatial bins, rapid

The max-pooled activations of
the rectified CNN filters coming from the last convolutional layer 
of the ZF-5 CNN network \cite{zeiler2014visualizing}
are another utilized proposal descriptor. 
This method, originally proposed by He \etal \cite{he2014spatial}, rapidly extracts features from spatial bins of several
thousand bounding boxes per second.
We $\ell_2$ normalize the CNN-SPP features to facilitate
the convergence of the later used SVM classifier.

%Since a SVM solver is used, normalization of the training data is a crucial step. Thus
%we further $\ell_2$ normalize the extracted CNN-SPP features.

%Recently, He \etal \cite{he2014spatial} have proposed a method 
%that is able to rapidly extract CNN descriptors from several thousands of
%regions per second. It consists of performing max-pooling of activations of
%the CNN filters coming from the last convolutional layer of the ZF-5 CNN network \cite{zeiler2014visualizing}
%inside individual spatial bins of each bounding box. We opted for using these
%pooled features as another proposal descriptor. Also, since a SVM
%solver is used, normalization of the training data is a crucial step. Thus
%we further $\ell_2$ normalize the extracted CNN-SPP features.

The layout of bounding box subdivisions is the same as in \cite{he2014spatial}, \ie the bounding box is split to $D^2$ equally sized divisions that cover a box uniformly without overlap (Figure \ref{fig:ibe-spabins}). We set multiple $D$ parameters such that 10 different split layouts are created corresponding to $D = \{1,2,3, ..., 10\}$, giving 385 bins in total. However, in practice \emph{we pool conv5 features only from the bins selected by the feature selection approach} which is thoroughly described in Section \ref{sect:binSparsity}.

%We discovered that a fine resolution of used spatial pyramid is important for achieving \sota performance of our objectness detector. Thus we max-pool conv5 features from 10 different split layouts corresponding to $D = {1,2,3, ..., 10}$, giving 385 bins in total.

\section{Boundary Edge Vector feature (BEV)}

\label{sect:bevfeat}

%The second type of descriptors exploits the edge map 
%computed during the acquisition of EdgeBoxes (\ie the output of 
%the Structured Edge Detector \cite{dollar2013structured}).
%In addition, the descriptors are again obtained 
%via pooling edge statistics inside individual bounding box spatial bins.

Boundary Edge Vector exploits the EdgeBoxes edge map (\ie the output of 
the Structured Edge Detector \cite{dollar2013structured})
for pooling edge statistics inside individual bounding box spatial bins.
More precisely, all edgels residing inside a spatial bin are quantized to 4 equally wide orientation bins. 
After that a 4-dimensional
bin descriptor is formed by utilizing integral images
to accumulate the edgel intensities that correspond
to each of the 
%edge 
orientation bins.
All these bin descriptors are then concatenated into a single vector which is later $\ell_2$ normalized to form the final BEV descriptor.

%All these 4-D features coming from every spatial bin of a given bounding box

%These 4-D spatial bin features 

\begin{figure}[t]
\begin{center}
\begin{tabular}{c}
   \includegraphics[width=0.65\linewidth]{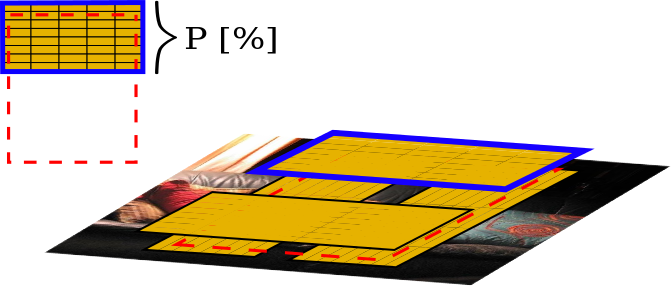} \\ (a) BEV layout \\
   \includegraphics[width=0.6\linewidth]{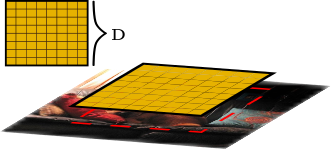} \\ (b) CNN-SPP layout
\end{tabular}
\end{center}
   \caption{\textbf{The layout of spatial bins used for pooling descriptors}. (a) BEV is pooled in 40 bins arranged
along each of the bounding box sides. (b) CNN-SPP descriptor spatial bins cover the bounding box uniformly without overlap.}
\label{fig:ibe-spabins}
\end{figure}

The layout of BEV spatial bins is depicted in Figure~\ref{fig:ibe-spabins}.
First, in order to include information about the 
%distribution of 
edges that
cross the bounding box boundary, {\changed the bounding box dimensions are both enlarged by 10\% prior to
creating the spatial subdivisions.}
Then, eight stripes collinear with each of the bounding box sides are all divided %"chopped"
across to five divisions to form 40 spatial bins per bounding box side in total.
The stripe octet's width is set, such that it covers $P \%$ of the bounding box side.
Several different layouts each corresponding to different
values of  $P$ \mbox{ ($P=\{0.16, 0.18, 0.22, 0.24, 0.28, 0.32, 0.36 \}$)} are used. Additionally, 
\emph{feature selection (explained in Section \ref{sect:binSparsity}) is again used to pick the most informative spatial
bins} and thus only the selected ones are again chosen for extraction of the bin descriptors.

%The final descriptor is a concatenation of all acquired bin descriptors.

%We pool the features in several different layouts each corresponding to different
%values of  $P$  ( {\red !!!!TODO $P = \{ ... \}$}). The final descriptor is a concatenation of all layout descriptors.

{\changed
The Boundary Edge Vector resembles the Boundary Edge
distribution (BE) proposed by Rahtu \etal in \cite{rahtu2011learning}.
However, in BE \cite{rahtu2011learning}, edgels corresponding to only one predefined edgel orientation bin are accumulated inside every spatial bin. 
Furthermore the accumulated orientation intensities are projected using a predefined set of weights whereas we "unfold" the descriptor into a much higher dimensional vector where all spatial orientations are taken into account. The SVM classifier determines the best weights for each orientation and spatial bin. Finally, we improve the pooling stage, by increasing the number of pooling bins and subsequently learning their optimal layout inside the spatial bin selection algorithm. In the light of these changes our newly introduced feature could be seen as a generalization of the Boundary Edge distribution measure.
}

% The Boundary Edge Vector bears resemblance to the Boundary Edge
% descriptor (BE) proposed by Rahtu \etal in \cite{rahtu2011learning}.
% However, in \cite{rahtu2011learning} a single number is the result of 
% the bounding box descriptor extraction. In each spatial bin only edgels corresponding to a
% predefined edgel orientation bin are accumulated. A weighted sum 
% of these 1D bin descriptors forms the final BE feature.
% %Thus each bin is characterized by 1D descriptor.
% %Instead of concatenation a weighted sum of these 1D features forms the final BEV feature. 
% In our case, 
% we "unfold" the descriptor
% and let the SVM classifier to determine the most convenient weights for
% each orientation and spatial bin.  

\section{Spatial bin selection}

\label{sect:binSparsity}

%We discovered that i
In the case of BEV and CNN-SPP features a large number of spatial bins 
has to be used in order to obtain \sota performance.
However, this substantially increases the computational demands.
We therefore perform a feature selection step
which automatically picks relevant spatial bins that will form the final descriptor.

Our descriptors are created by pooling information from spatial bins, they are formed
by {\changed groups} of values that correspond to spatial subdivisions.
To perform selection of bins we use a sparsity-inducing SVM solver \cite{yang2010online}, that 
employs the \emph{group lasso} term as a regularizer $\Omega(w)$ \cite{yuan2006model}. 
More precisely $\Omega(w) = \sum_{b=1}^{B} \|w_b\| $ where $w$ stands
for the set of SVM weights, $w_b$ is the group of weights corresponding to
the bin $b$ and $B$ is the overall number of used subdivisions.
{\changed The value of the $C$ parameter controls the number of zeroed groups~$w_b$.} 
%By varying the solver's $C$ parameter, one can induce sparsity on the level of individual groups of dimensions of the learned SVM weights. 

Each spatial bin that corresponds to a group of zeroed SVM weights then plays no role
in the final bounding box score and thus could be omitted during the
feature extraction step. {\changed For BEV and CNN-SPP descriptors the groups of dimensions have size 4 (number of orientation bins) and 256 (number of convolutional filters) respectively.
}
%The used group sparsity solver comes from \cite{yang2010online}. 

% {\red COMPRESS}
{\changed
Our choice of group normalized SVM, instead of e.g. $\ell_1$ regularized SVM which would remove individual descriptor dimensions, is motivated by the computational overheads associated with visiting a single spatial bin:
for conv5 features, the spatial bin max-pooling is implemented using SSE instructions thus it is faster to access one continuous block of memory, represented by all convolutional features inside a spatial bin. For BEV, memory addresses to the integral image have to be computed. 
Thus, by using group normalization, we not only avoid computation of many features but we also decrease the number of costly visits of spatial bins.
}

Note that the approach consisting of 
inducing block sparsity to image features was first used in \cite{novotny2015understanding}
to discover relevant gaussians in the context of Fisher Vector detection pipeline \cite{cinbis2013segmentation,perronnin2010improving}.

% {\changed
% Note that in our case, removing individual descriptor dimensions by using e.g. $\ell_1$ regularized SVM may sound more appropriate than removing the whole groups of dimensions. However, we discovered that processing a whole spatial bin (i.e. max-pooling over all the conv5 filters and accumulating edgel intensities from every orientation bin) has roughly the same expense as extracting only a partial information from it. In the case of conv5 features, the max-pooling is implemented using SSE instructions. It is thus faster to access one continuous block of memory, represented by all the convolutional filters than using a naive implementation for pooling individual filters. For BEV features, there are overheads associated with visiting a single spatial bin (\eg computation of the 4 memory addresses in the integral image). Thus it is more efficient to extract all the information once a particular bin is entered.
% }

\section{SVM and group normalized SVM learning}

\label{sect:svmLearning}

The standard SVM that combines BEV, CNN-SPP and EB features as well as the group lasso SVM classifiers are learned on the same set of training bounding boxes. The positive examples are all the ground truth regions that contain any of the object classes present in the "train"+"val" sets of the Pascal VOC 2007 detection dataset \cite{everingham2010pascal}. 

The set of negative bounding boxes is composed of two equally sized subsets. While all regions are required to have at most 30\% overlap (Pascal intersection-over-union metric) with any of the ground truth objects the first half
is sampled from the immediate vicinity of the ground truth regions, while boxes from the second can reside at any location in any training image.
%The size of the number of negative samples is one half of 
%There is twice as many positive samples as negatives.
The number of negative samples is roughly equal to half of the positive samples.

%Following \cite{novotny}, the sparsified dimensions are removed from the training descriptors
%and a standard $\ell_2$ normalized SVM classifier is retrained 
%on the set of descriptor dimensions selected by the group normalized SVM.

% The first contains randomly created regions from the immediate vicinity of training ground truth bounding boxes but do not overlap them by more than 30\% (Pascal intersection-over-union metric). The second part consists of regions sampled at random from any part of any training image, while they are again restricted to have at most 30\% overlap with any of the ground truth objects. The number of negative samples is roughly equal to half of the positive samples.

After the three aforementioned descriptors are obtained from each training region,
the sparsity-inducing learning follows. Since the sizes of groups of dimensions that
we want to remove are distinct for each of the two feature types (BEV and CNN-SPP),
we train two different sparse SVM classifiers separately for each descriptor design.
In practice, for the second stage of the detection cascade we select the SVM's regularization parameter such that 43 and 311 spatial regions are selected for CNN-SPP and BEV features respectively. In the case of the first stage of the SSPB classifier, which utilizes CNN-SPP feature, only 3 spatial bins were selected. %These parameters were optimized on the Pascal VOC2007 validation set and were found to provide the best compromise between the speed and the accuracy of the whole detection system.

After the feature selection step, following \cite{novotny2015understanding}, training descriptors are stripped of the unused dimensions and $\ell_2$ re-normalized. Additionally, the survivors of the feature selection process are concatenated and the corresponding EB feature is appended to form the final set of training descriptors for the standard $\ell_2$ regularized SVM learning. The $C$ regularization parameter of the $\ell_2$ regularized SVM was set to 1. Hard negative mining tends to worsen the detector performance. We thus stop the pipeline training after the initial mining of random negative samples.

\section{Non-maximum suppression for optimizing average recall}

\label{sect:arnms}

We discovered that it is suboptimal to perform the standard greedy NMS for discarding redundant high scoring regions, since it tends to either remove many well-located proposals, when its threshold is set to a low value, or completely miss a large portion of regions that are not finely aligned with an object (when the NMS threshold is set to a high value). 

To reach a compromise between these two situations, we employ a special type of NMS which we term ARNMS. The goal of ARNMS is to extract a set of candidates that have the best possible average recall given the desired number of output object detections $N$. More accurately, ARNMS runs in $S$ subsequent stages. During stage $s$ the standard greedy NMS is performed with overlap threshold $o_s$ followed by the extraction of $N/S$ highest scoring not suppressed regions. In practice we use $S = 3$ with $o_1 = 1$ (\ie no NMS employed), $o_2 = 0.7$ and $o_3 = 0.5$. %We observed that ARNMS is able to improve recalls levels at both high and low IoU thresholds. 

Note that \cite{hosang2015what} have employed a similar strategy for improving the AR of EdgeBoxes proposals. However their approach is not tuned for specific amounts of outputted proposals thus, when for instance using a very dense sampling of EdgeBoxes during the first stage of our SSPB cascade, the method of \cite{hosang2015what} would be comparable to running simple greedy NMS with threshold set to 0.9 resulting in low recalls at decreased IoU thresholds.

\section{Experiments}

\label{sect:experiments}

We test our object proposal methods on two standard object detection benchmarks:

\textit{VOC07 \cite{everingham2010pascal}}: The "test" and in some cases "val" sets of the Pascal VOC2007 dataset were used for evaluation of our methods. The "test" set (VOC07-TEST) consists of 4952 images containing 20 distinct visual object classes together with their bounding box annotations. 2510 images are included in the "val" set (VOC07-VAL) and a similar number of 2501 pictures resides in the "train" set (VOC07-TRAIN).

\textit{ILSVRC2013 \cite{ILSVRCarxiv14}}: To check the ability of our method to generalize to unseen data the more challenging ILSVRC2013 DET task's "validation" set was utilized. The amount of images is roughly 20k while there are annotations for 200 object classes.  

Note that abbreviations of all competing proposal methods are matched to their original papers in the References section.

%, we also concluded experiments on 

% Since our detection system was trained on the "train" and "val" sets of the Pascal VOC2007 \cite{everingham2010pascal} (abbreviated as VOC07) our experiments were concluded on the "test" and some one the "val" subsets of VOC07. Additionally we also present results on the "validation" set of ILSVRC2013 \cite{ILSVRCarxiv14} dataset.

\begin{figure*}[t]
\begin{center}
\begin{tabular} {c c c} 
\includegraphics[width=0.295\linewidth]{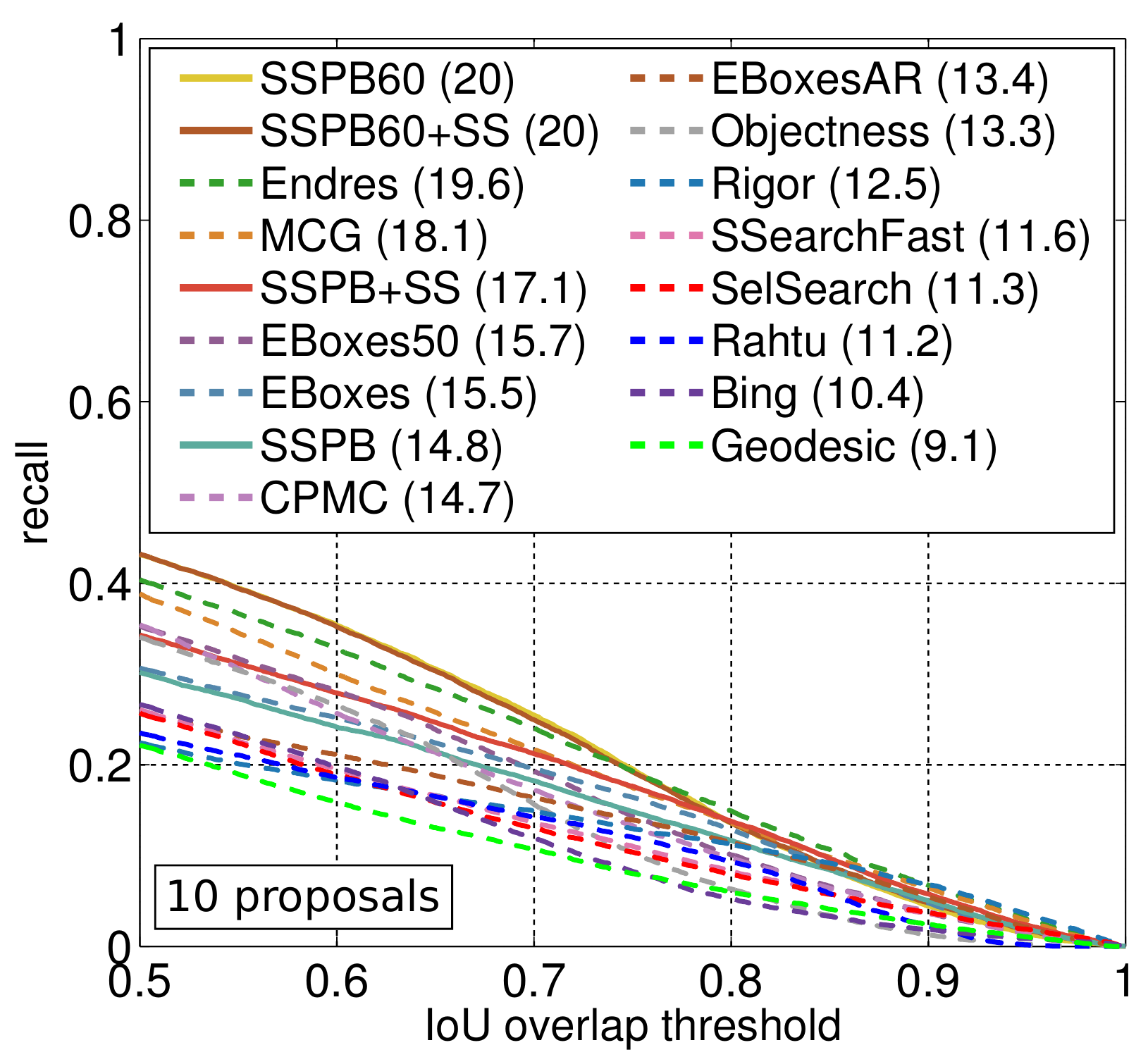} &
\includegraphics[width=0.34\linewidth]{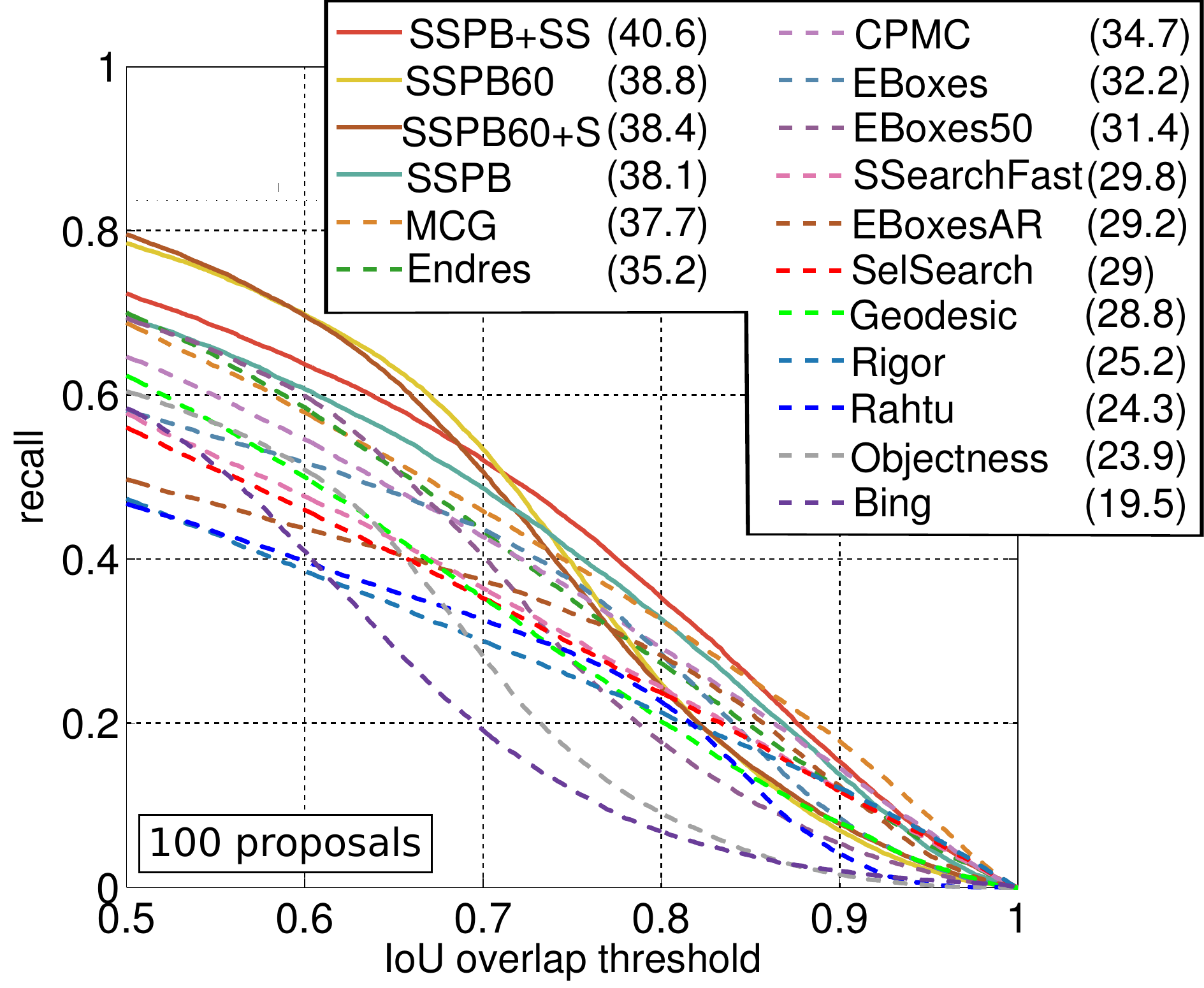} &
\includegraphics[width=0.345\linewidth]{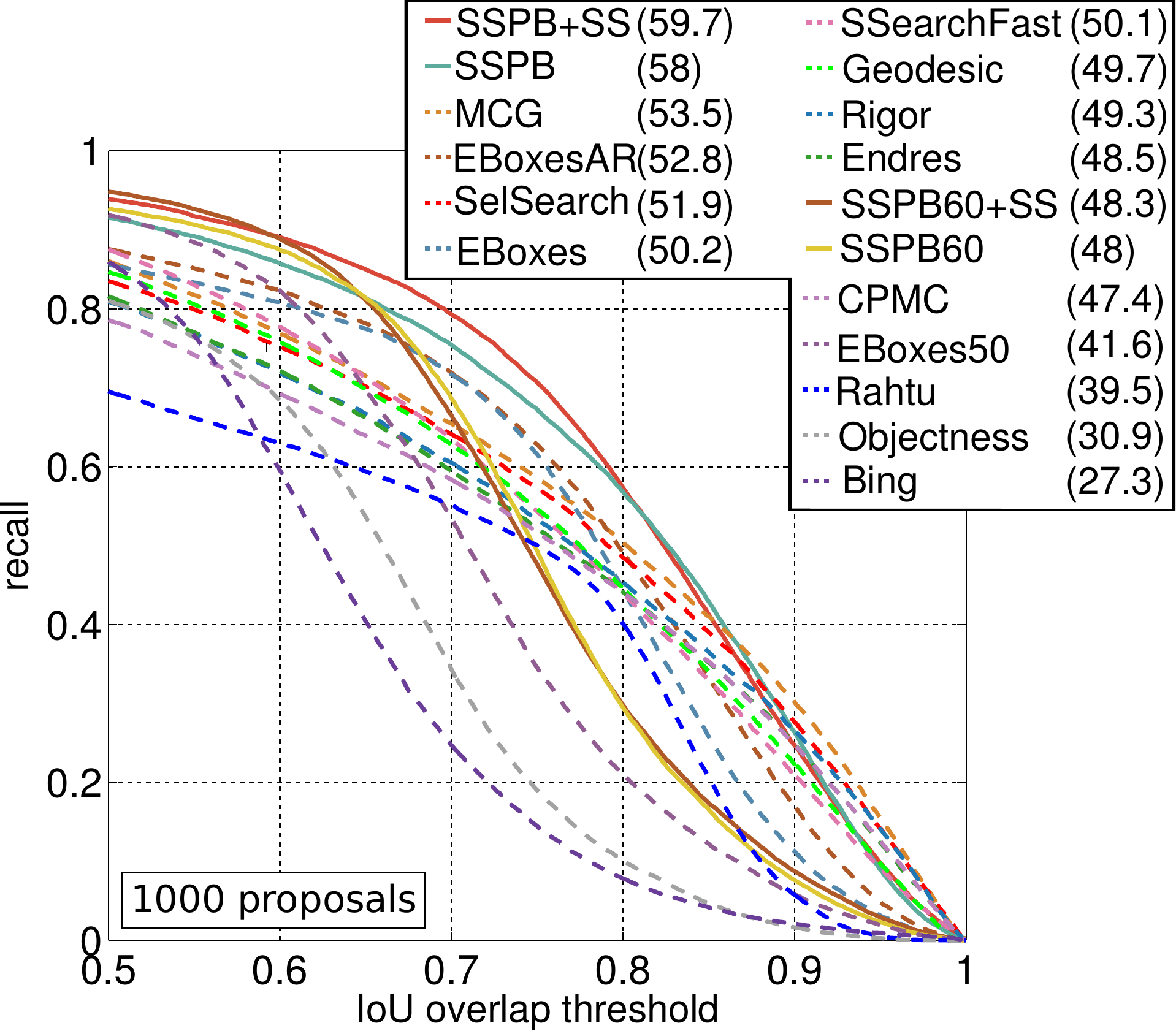} \\
\end{tabular}
\end{center}
   \caption{\textbf{Overlap-recall curves} of our (solid lines) and \sota proposal methods on the VOC07-TEST set when 10~(left), 100~(center) and 1000~(right) candidate windows are considered per image. The legends are sorted by the average recalls (in brackets).}
\label{fig:overlap-recall-voc07}
\end{figure*}

%\subsection{Overlap-recall experiments}
\noindent
{\bf Overlap-recall experiments}
%In this section, the we report achieved recalls as a function of the minimal requires IoU overlaps obtained using the publicly available benchmark code made by Hosang \etal \cite{hosang2014good}
%
In this section, overlap-recall curves obtained using the publicly available benchmark code made by Hosang~\etal~\cite{hosang2014good}, are provided.
In case of overlap-recall curves an "oracle" detector that for each ground truth bounding box reports the most overlapping proposal is run. 
The curve then consists of achieved recalls
as a function of minimal required IoU overlaps at which a proposal is regarded as a true positive.

%The first group of experiments evaluates achieved recalls while changing minimal required IoU overlaps at which a proposal is regarded as a true positive. {\red REMOVE? Furthermore, in the case of evaluation of the proposal methods, an "oracle" detector is typically employed, which for each ground truth bounding box always reports the most overlapping region from the produced pool of proposals.}
%Results in this section utilize the publicly available benchmark code made by Hosang \etal \cite{hosang2014good}.

We tested 4 variants of our algorithm. SSPB and SSPB+SS (described in detail in the preceding sections), SSPB60 and SSPB+SS60.
SSPB60 differs from SSPB in the final step where ARNMS is replaced by the standard greedy NMS with the overlap threshold set to 0.6.
The same applies to SSPB+SS60, which replaces the ARNMS step of SSPB+SS. The two additional methods were introduced because they give compelling performance when a small amount of candidates is requested.

%The only difference between SSPB60 and SSPB is 
%that the final ARNMS step of SSPB is replaced by the standard NMS with the overlap threshold set to 0.6. 

\begin{figure}
\begin{center}
\includegraphics[width=0.8\linewidth]{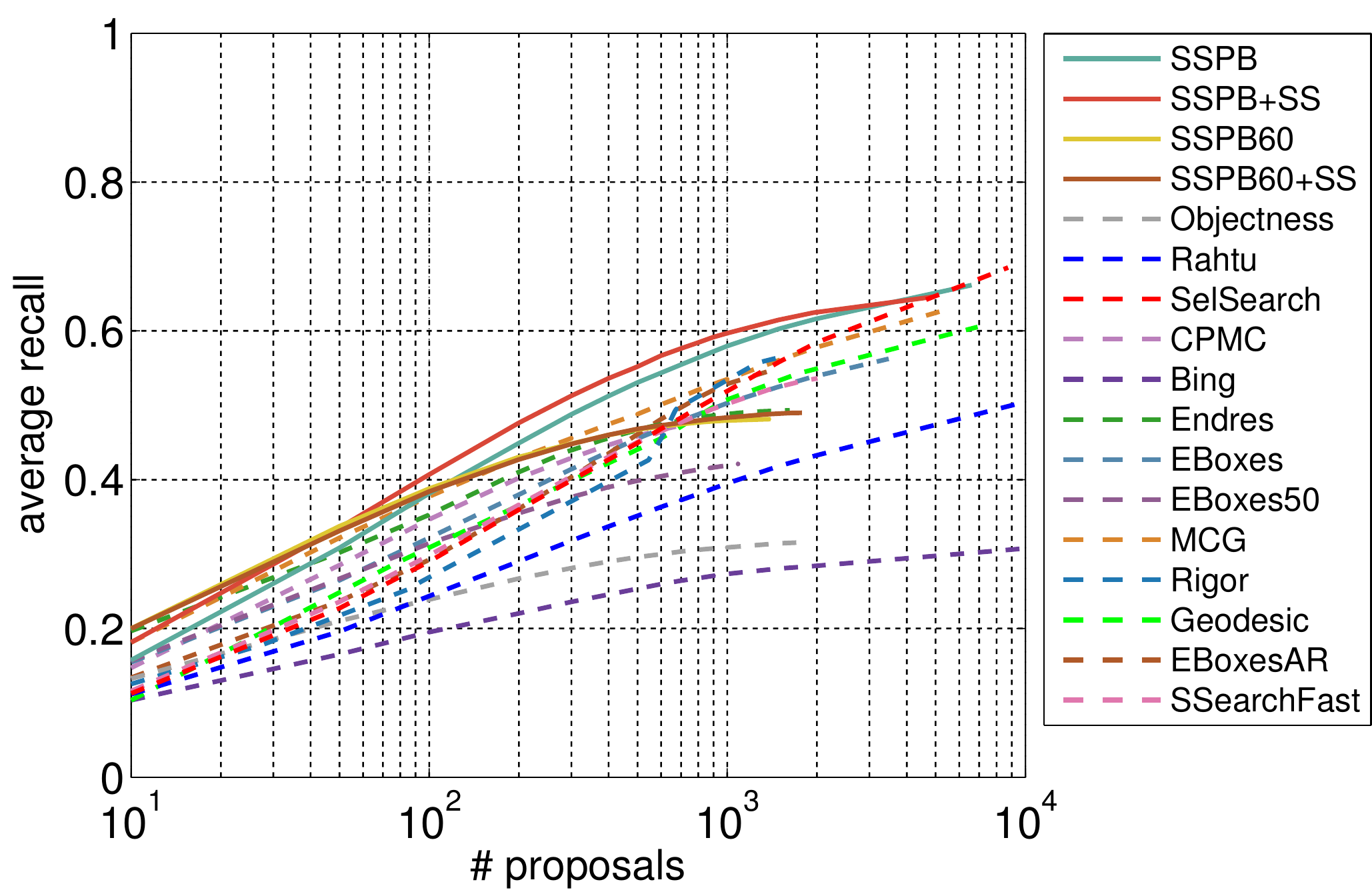} 
\end{center}
\caption{\textbf{Average recalls achieved by our (solid lines) and \sota proposal approaches on VOC07-TEST} as a function of the number of proposals per image.}
\label{fig:ar}
\end{figure}

Figure \ref{fig:overlap-recall-voc07} shows the overlap-recall curves of our methods on VOC07-TEST in comparison with \sota algorithms. Additionally, in Figure \ref{fig:ar} we provide average recall measures that have been shown to conveniently quantify the performance of generic object detectors \cite{hosang2015what}.

It is apparent that our approach performs better or on par with \sota both in terms of average recall and the individual recalls achieved at most IoU thresholds. It is rivaled only by the Selective Search ("quality mode") when 10000 candidate windows per image are considered. As noted earlier SSPB and SSPB+SS do not give that impressive performance when a small number of candidates is requested, however the decrease of the non-maximum suppression threshold (SSPB+SS60 and SSPB60) puts our approaches again in the leading position.
The comparison between SSPB and SSPB+SS is slightly in favor of SSPB+SS, however we note that SSPB is faster due to the skipping of the Selective Search extraction step. Another positive point is that although SSPB is categorized as one of window scoring methods that tend to attain lower recalls at higher IoU thresholds, it is able to produce bounding boxes comparable to those of \eg MCG or Selective Search in terms of localization quality.

%\subsection{Combination with a class-specific detector}
\noindent
{\bf Combination with a class-specific detector.} To check the applicability of our method, we designed an experiment where the \sota RCNN \cite{girshick2014rich} class-specific object detector utilizes the output of a proposal generation algorithm. The four proposal algorithms that were tested were SSPB, SSPB+SS, EdgeBoxes70 and Selective Search in its "fast mode" (originally used for RCNN). We recorded the achieved RCNN mAPs on the VOC07-TEST set while varying the number of used candidates per window.

Since we empirically discovered that using a proper IoU threshold when executing the non-maximum suppression of SSPB, SSPB+SS and EdgeBox70 candidate windows is crucial for obtaining the best possible final RCNN performance, we validated these optimal NMS thresholds on the VOC07-VAL set for each number of requested proposals separately. Table \ref{tab:rcnn} shows achieved mAP values.

%contains precise values of achieved mAP, while Figure  captures the same information in a more transparent form.

The results indicate that in case of SSPB and SSPB+SS the RCNN mAP decreases the least as the number of candidates is reduced. Also note that RCNN was originally trained using the Selective Seach "fast" proposals, which typically sways the results in favor of this method  \cite{hosang2015what}.%used during the class-specific classifier training \cite{hosang2015what}. 
Yet, SSPB+SS and SSPB is still able to outperform Selective Search "fast"; additionally our methods improve the results of the original RCNN pipeline when 1000 and more proposals are produced per image.

\begin{table}
\begin{center}
\footnotesize
\begin{tabular}{|c|c|c|c|c|c|c|}
\hline
& \multicolumn{6}{ |c| }{\# candidates} \\ \hline
method &10 &50 &100 &500 &1000 &10000\\ \hline
%SSFast & 23.7 &37.2 &42.8 &52.5 &54.2 &54.8\\
%EB&32.3 &43.0 &46.1 &52.1 &53.3 &53.1\\
%SSPS (ours)&\textbf{37.6} &47.3 &\textbf{50.3} &53.2 &56.0 &56.0\\
%SSPB+SS (ours)&35.7 &\textbf{47.8} &50.2 &\textbf{56.1} &\textbf{56.6} &\textbf{56.3}\\
SSFast&23.7 &37.2 &42.8 &52.5 &54.2 &54.8\\
EB      &32.3 &43.0 &46.1 &52.1 &53.3 &53.1\\
SSPS (ours)&\textbf{36.0} &46.7 &50.0 &53.1 &56.4 &\textbf{56.3}\\
SSPB+SS (ours)&35.7 &\textbf{47.8} &\textbf{50.2} &\textbf{56.1} &\textbf{56.6} &\textbf{56.3}\\
DMultiBox \cite{erhan2014scalable}&29.0& - & - & - & - & - \\ \hline
\end{tabular}
\end{center}
\caption{\textbf{RCNN detector mAP as a function numbers of proposals per image} for different proposal methods.}
\label{tab:rcnn}
\end{table}

Recently DeepMultiBox \cite{erhan2014scalable} attained 29.0 mAP on VOC07-TEST with their proposals and using a class-specific detector with CNN architecture from \cite{krizhevsky2012imagenet} while considering just 10 candidate regions per image.
Our result is substantially higher, achieving 36.0 mAP when the same number of SSPB regions is proposed in each image while noting that the architecture of RCNN differs from DeepMultiBox only in the type of classifier in the topmost layer.

%While noting that {\red the only difference in architecture between the class-specific detector from \cite{erhan2014scalable} and RCNN is that in the latter case SVM classifiers are utilized instead of the topmost softmax layer}, our result is substantially higher, achieving 37.6 mAP when the same number of SSPB regions is proposed in each image. 

%\subsection{Generalization experiments}
\noindent
{\bf Generalization experiments}
Since our method is trained on VOC07-TRAIN+VOC07-VAL, which contains only 20 distinct classes, we verified its performance in a more challenging setting with previously unseen classes. We thus tested the proposed method on the ILSVRC2013 validation set of the detection task.

\begin{figure*}[t]
\begin{center}
\begin{tabular} {c c c}
\includegraphics[width=0.30\linewidth]{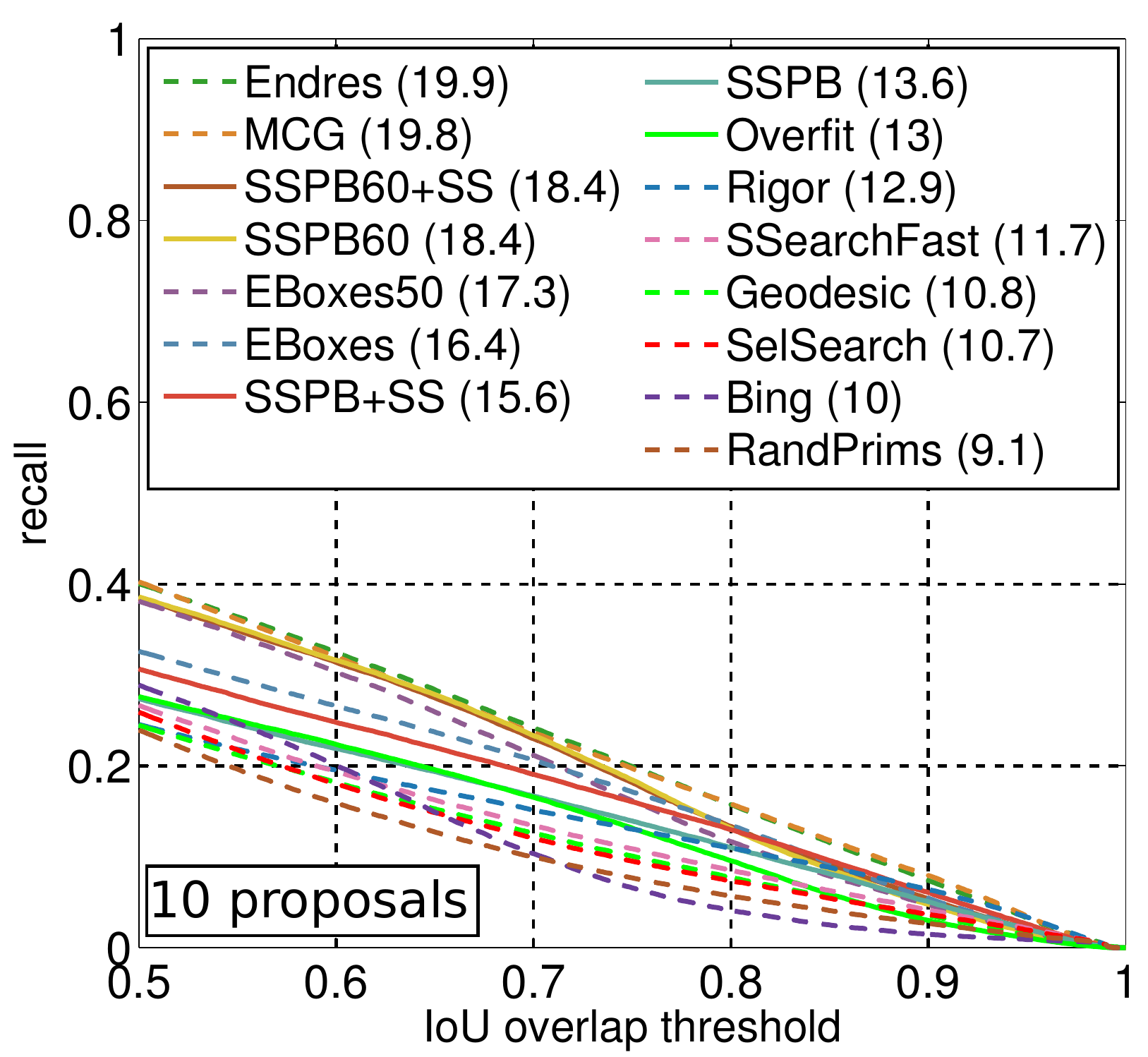} &
\includegraphics[width=0.29\linewidth]{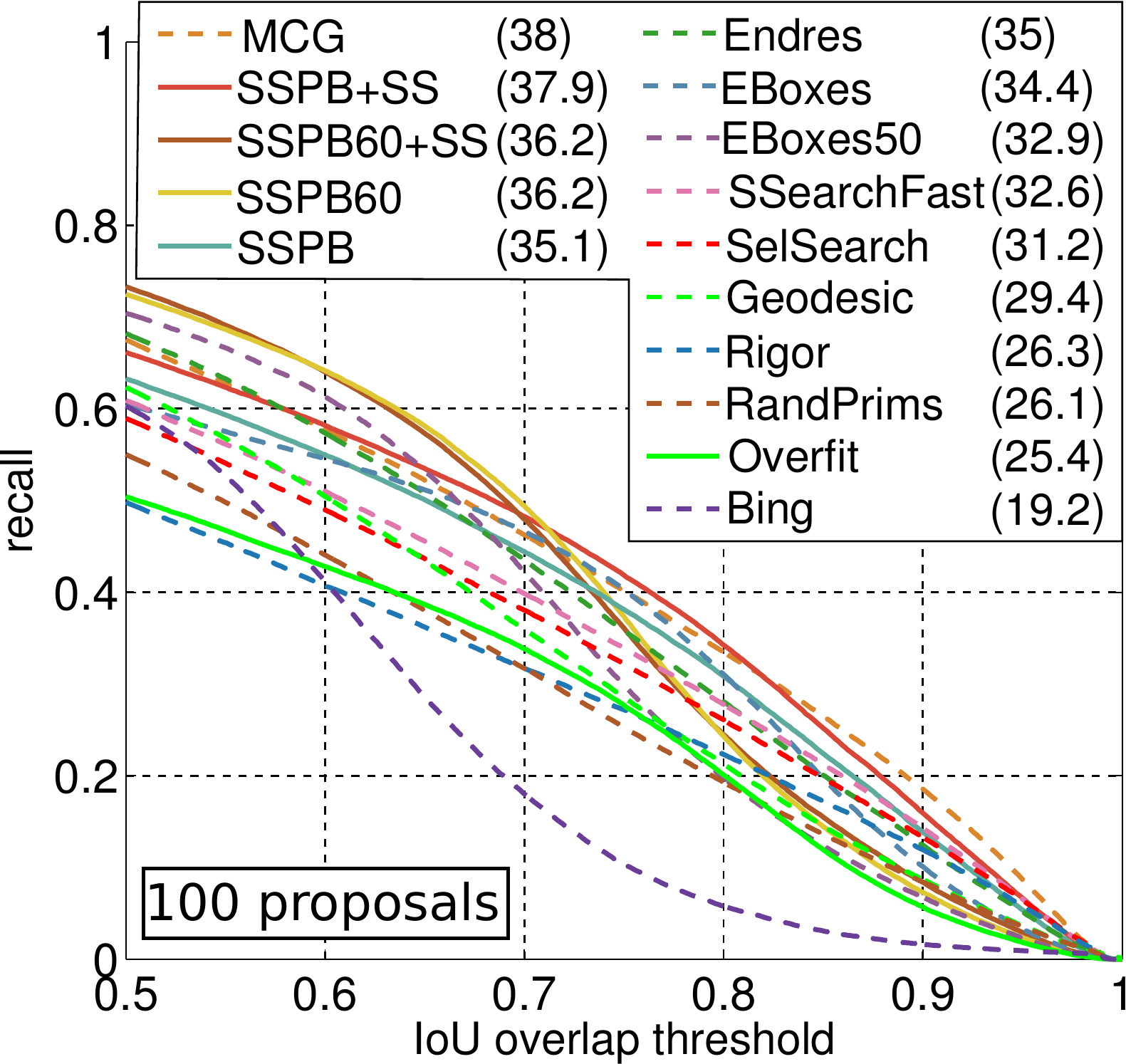} &
\includegraphics[width=0.352\linewidth]{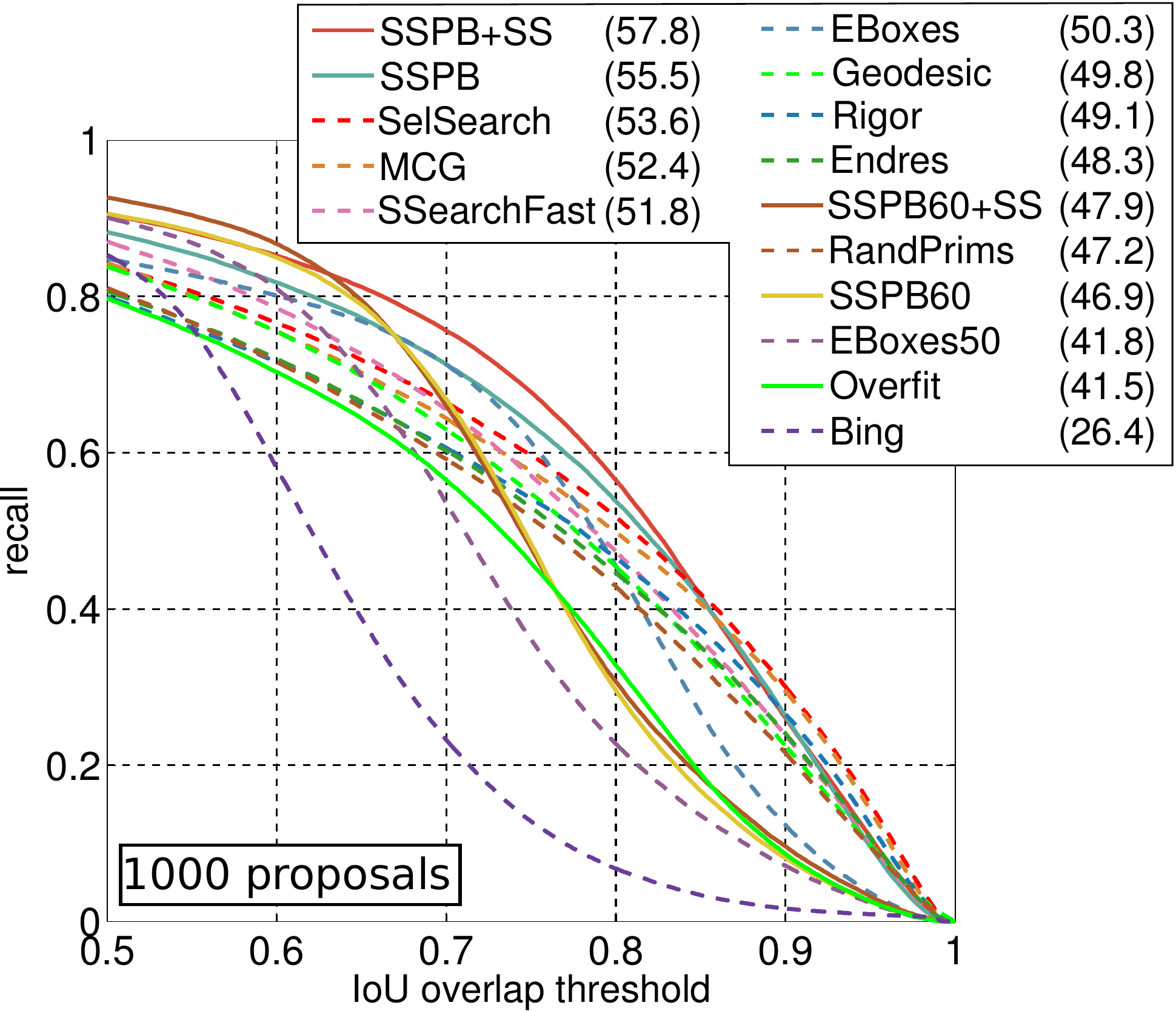} \\
\end{tabular}
\end{center}
   \caption{\textbf{Overlap-recall curves} of our (solid lines) and \sota proposal methods on \mbox{ILSVRC2013-DET-VAL$_R$} set when 10 (left), 100 (center) and 1000 (right) candidate windows are considered per image. The legends are sorted by average recalls (in brackets).}
\label{fig:overlap-recall-ilsvrc2013}
\end{figure*}

A potential caveat is that the used ZF-5 network was trained on the ILSVRC2013 training set of the classification task which contains some of the images from the ILSVRC2013 validation set of the detection task used for testing our detector. To overcome this problem we removed the 311 images that are located in both sets and tested our detector on this very slightly reduced set (we refer to it as \mbox{ILSVRC2013-DET-VAL$_R$}). Overlap-recall curves  of our and \sota proposal techniques are plotted in Figure \ref{fig:overlap-recall-ilsvrc2013}, again with the help of the software of \cite{hosang2015what}.

{\changed Recently, \cite{chavali2015gameable} have shown that the object proposal evaluation protocol could be gamed by training a class specific object detector and use its scores as an objectness measure. We trained an objectness SVM classifier on the 20 class Pascal scores produced by the original CNN-SPP detector \cite{he2014spatial} according to the training protocol from Section \ref{sect:svmLearning}. We then apply this classifier to \mbox{ILSVRC2013-DET-VAL$_R$}. The method is labeled "Overfit" in Figure \ref{fig:overlap-recall-ilsvrc2013}.}

Results show that our approaches outperform other methods in terms of AR as well as in achieved recalls evaluated between 0.5 and 0.83 overlap thresholds, once a larger amount of candidate windows is considered (more than 500). For the lower proposal amounts SSPB and SSPB+SS stay on par with the competition. {\changed Our methods also outperform the Overfit proposals.}

%\subsection{Feature selection, run-time analysis}
\noindent
\textbf{Feature selection experiments} demonstrate the ability of the employed feature selection algorithm to decrease the number of spatial bins while maintaining comparable performance of our approach. We trained two different variations of SSPB+SS on VOC07-TRAIN set and tested them on VOC07-VAL. The first solely utilized the CNN-SPP descriptors, whilst the second used only BEV descriptors. Average recalls as a function of the number of used proposals for various numbers of selected spatial bins are reported.

\begin{figure}
\begin{center}
\includegraphics[width=0.6\linewidth]{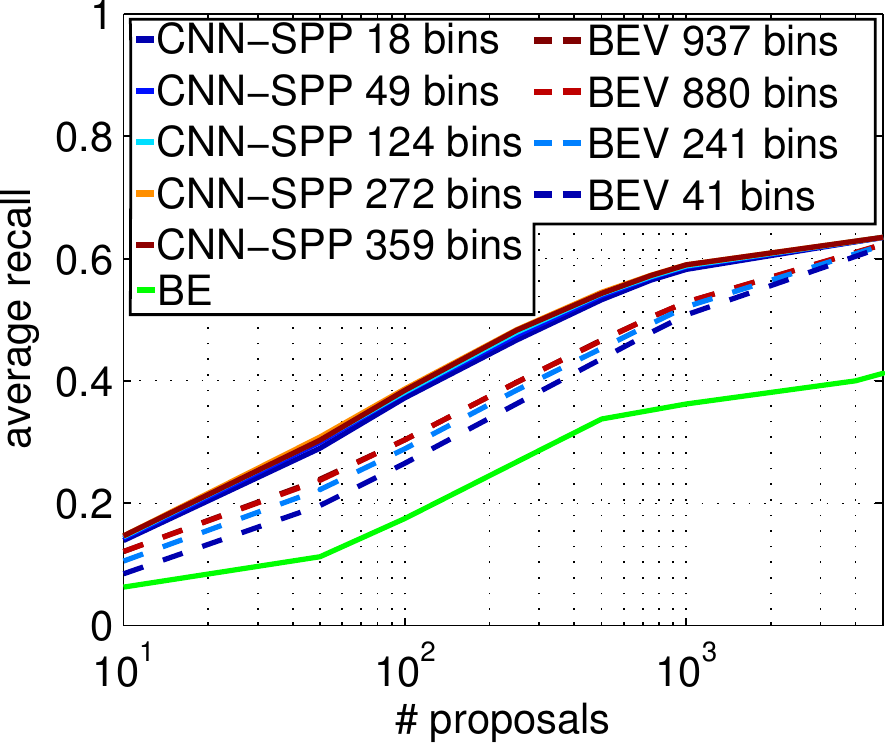}
\end{center}
\caption{\textbf{Average recalls achieved by BEV and CNN-SPP features respectively as a function of the number of proposals} for different numbers of spatial bins selected by group-lasso SVM. {\changed The BE feature \cite{rahtu2011learning} is included to show its inferiority to BEV.} }
\label{fig:featSel}
\end{figure}

Figure \ref{fig:featSel} shows that for CNN-SPP as well as for BEV features, the resulting average recall decreases very slowly as the number of effective spatial bins is reduced. This way it is possible to limit the amount of utilized spatial subdivisions to 25 \% and 85 \% of the original number in the case of BEV and CNN-SPP features respectively, without hurting the quality of the produced proposals. Figure \ref{fig:featSel} contains performance of the BE feature \cite{rahtu2011learning} to show the improvement of the BEV feature over BE.

%\subsection{Run-time analysis}
\noindent
{\bf Run-time analysis.}  The speed of our methods is compared with the algorithms that attained the best average recalls in experiments. Table \ref{tab:times} shows mean processing times on a fixed subset of 200 images sampled randomly from VOC07-TEST. {\changed We report both GPU (GeForce GTX TITAN BLACK) and CPU (Intel Xeon E5-2630~v3) times.

% For extraction of conv5 feature maps we used CPU which is not in agreement with common practice of utilizing a GPU due to the fact that it drastically slows down the whole detection pipeline. Thus as conv5 extraction times we include the values provided in \cite{he2014spatial} where a GPU was used  for the computation of the conv5 features identical to those we use.

The speed of our approach is comparable to the Selective Search "fast mode" and roughly 4$\times$ faster than its "quality mode", while performing better than the "quality mode". Most of the segmentation methods that perform on par with SSPB and SSPB+SS on ILSVRC2013 and are inferior on VOC07-TEST such as MCG, Endres or CPMC are slower by more than an order of magnitude. 
}

\begin{figure}
\begin{center}
\includegraphics[width=0.8\linewidth]{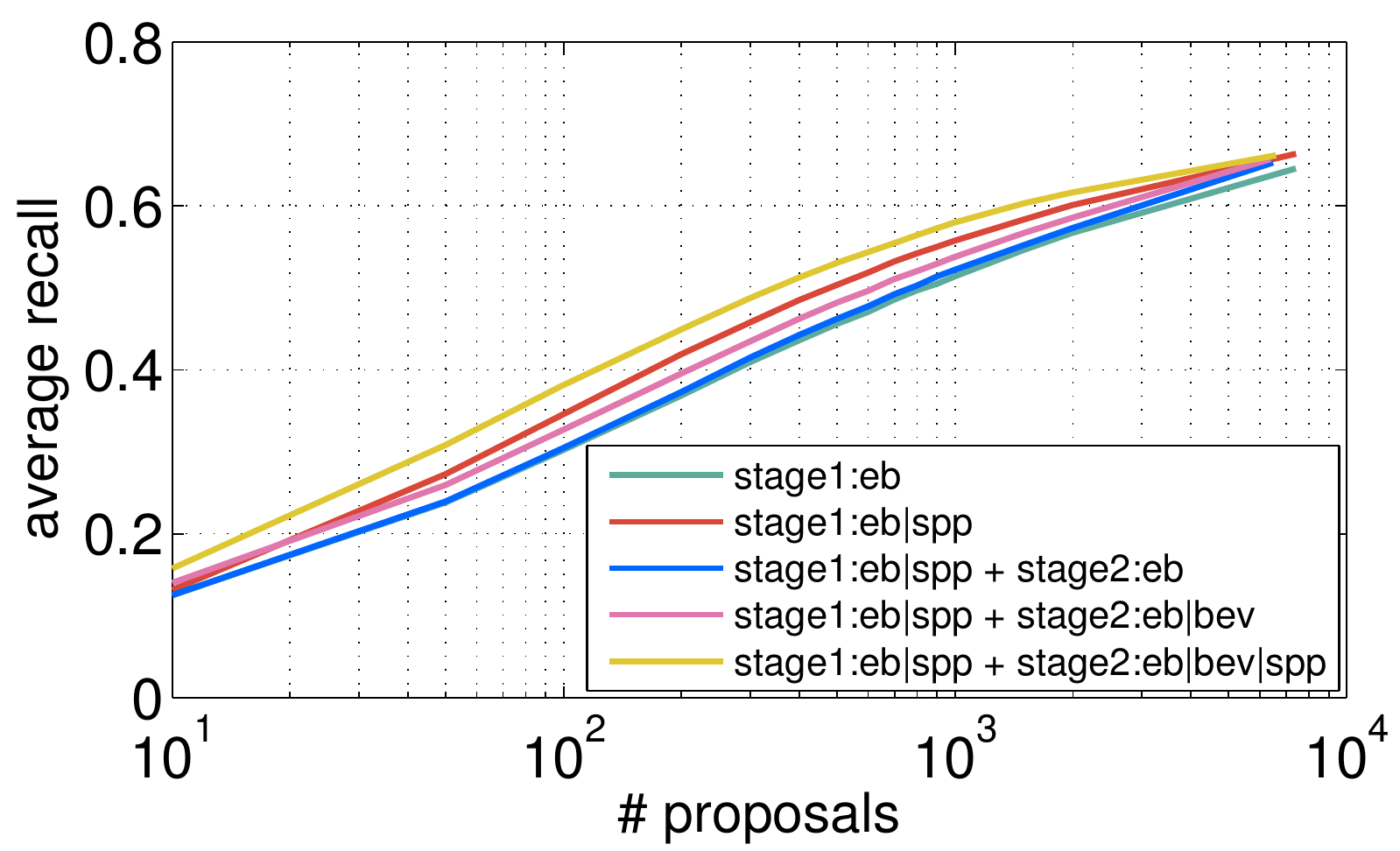}
\end{center}
\caption{{\changed\textbf{Ablation analysis.} Performance on VOC07-TEST of both stages of SSPB with different feature combinations.}}
% \caption{\textbf{mAP attained by the RCNN detector when used in combination with 4 different proposal methods.} The "x" axis contains the number of used proposals per image, the "y" axis depicts the mAP of the RCNN detector.}
\label{fig:rcnn}
\end{figure}

\noindent
{\changed \textbf{Ablation analysis.} Figure \ref{fig:rcnn} presents average recall curves achieved on VOC07-TEST with each stage of SSPB while varying the combination of proposed features. Results show beneficial effects of each added feature/stage.
}

%\begin{itemize}
%\item Fast as SSFast, but better
%\item Much faster than segmentation based methods (MCG, CPMC, Endres), while better on Pascal and on par on ILSVRC
%\item window scoring approach with good localization
%\end{itemize}

%most of the \sota methods in terms of both speed and attained recall levels. 

\begin{table}
\begin{center}
\small
\begin{tabular}{|c|r||c|r|}
\hline
Method & time [s] & Method & time [s] \\ \hline
SSFast \cite{van2011segmentation} & 2.52  & MCG\textsuperscript{\textdagger} \cite{arbelaez2014multiscale} & 30 \\
SSQuality \cite{van2011segmentation} & 11.85 & CPMC\textsuperscript{\textdagger} \cite{carreira2010constrained} & 250 \\
EdgeBoxes70 \cite{zitnick2014edge} & 0.39 & Endres\textsuperscript{\textdagger} \cite{endres2010category} & 100 \\
SSPB (ours) & 3.16 (11.51) & Rigor\textsuperscript{\textdagger} \cite{humayun2014rigor} & 10  \\
SSPB+SS (ours) & 4.09 (12.45) & Geodesic\textsuperscript{\textdagger} \cite{krahenbuhl2014geodesic} & 1\\
\hline
%\begin{tabular}{|c|c|c|c|}
%\hline
%SSFast \cite{van2011segmentation} & SSQuality \cite{van2011segmentation} & EdgeBoxes70 \cite{zitnick2014edge} & SSPB (ours)  \\
%2.52 & 11.85 & 0.39 & 2.67  \\  \hline
%MCG \cite{arbelaez2014multiscale} & CPMC \cite{carreira2010constrained} & Endres \cite{endres2010category} & SSPB+SS (ours) \\
%30 & 250 & 100 & 3.61 \\ \hline
\end{tabular} \\
\hspace{5cm}{\footnotesize \textsuperscript{\textdagger}results taken from \cite{hosang2015what}}
\end{center}

\caption{\textbf{Per image processing times} of our methods and the best performing competition. {\changed For SSPB and SSPB+SS, GPU was used for extraction of conv5 features. Full-CPU times are in brackets.} }
\label{tab:times}
\end{table}

\section{Conclusions}

We have introduced a novel window scoring method for extraction of object proposals, named SSPB. SSPB uses several fast-to-compute features: 
deep CNN-SPP features, the EdgeBoxes score and the newly proposed BEV descriptor that accumulates information about edges near bounding box boundaries. 
We substantially speed up the extraction of these objectness cues by a group normalized SVM based feature selection which does not hurt the final generic object detector performance. The improvement decreased the SSPB processing times below the level of the majority of \sota proposal approaches. 

Results on the Pascal VOC2007 dataset indicate that our method delivers \sota  in average recalls and at recall levels at many IoU thresholds for various numbers of candidate windows per image.  We obtained similar results on ILSVRC2013.  Since SSPB was trained on Pascal, the positive results prove that  our method generalizes to previously unseen data.

% which, in combination with the fact that SSPB was trained on Pascal, means that
Our proposals work very well in combination with the current \sota class-specific object detector RCNN \cite{girshick2014rich}. Besides significantly improving RCNN mAP when the number of considered candidates is limited, higher numbers of SSPB proposals also slightly increase class-specific detection above the level of Selective Search.

%the fact that RCNN attains better mAP when using our candidates once a very small amount of them is used per image, SSPB proposals are also able to slightly improve the resulting mAP once higher amounts of candidates are requested. 

\label{sect:conclusion}

{\small
\bibliographystyle{ieee}
\bibliography{biblio}
}

\end{document}